\documentclass{article}

\usepackage{arxiv}

\usepackage[utf8]{inputenc} 
\usepackage[T1]{fontenc}    
\usepackage[hidelinks]{hyperref}       
\usepackage{url}            
\usepackage{booktabs}       
\usepackage{amsfonts}       
\usepackage{nicefrac}       
\usepackage{microtype}      
\usepackage{lipsum}         
\usepackage{graphicx}
\usepackage{natbib}
\usepackage{doi}
\usepackage{nicefrac}
\usepackage{amsthm}
\usepackage[acronym]{glossaries}
\usepackage{subcaption}
\usepackage{booktabs} 
\usepackage{multirow}
\usepackage{fontawesome}
\usepackage{tikzit}
\usepackage{cleveref}       
\usepackage{wrapfig}
\usepackage{tcolorbox}
\usepackage{xcolor}

\usepackage{algorithm}
\usepackage{algorithmic}

\title{Multi-Source Domain Adaptation meets Dataset Distillation through Dataset Dictionary Learning}

\date{}

\newif\ifuniqueAffiliation
\uniqueAffiliationtrue

\ifuniqueAffiliation 
\author{
Eduardo Fernandes Montesuma\\
CEA, List\\
Université Paris-Saclay\\
F-91120 Palaiseau, France
\And
Fred Ngolè Mboula\\
CEA, List\\
Université Paris-Saclay\\
F-91120 Palaiseau, France
\And
Antoine Souloumiac\\
CEA, List\\
Université Paris-Saclay\\
F-91120 Palaiseau, France
}
\else
\usepackage{authblk}

\newbox{\orcid}\sbox{\orcid}{\includegraphics[scale=0.06]{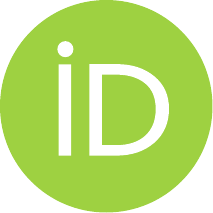}} 
\author[1]{%
	\href{https://orcid.org/0000-0000-0000-0000}{\usebox{\orcid}\hspace{1mm}David S.~Hippocampus\thanks{\texttt{hippo@cs.cranberry-lemon.edu}}}%
}
\author[1,2]{%
	\href{https://orcid.org/0000-0000-0000-0000}{\usebox{\orcid}\hspace{1mm}Elias D.~Striatum\thanks{\texttt{stariate@ee.mount-sheikh.edu}}}%
}
\affil[1]{Department of Computer Science, Cranberry-Lemon University, Pittsburgh, PA 15213}
\affil[2]{Department of Electrical Engineering, Mount-Sheikh University, Santa Narimana, Levand}
\fi

\renewcommand{\shorttitle}{Dataset Dictionary Learning}

\hypersetup{
pdftitle={Multi-Source Domain Adaptation meets Dataset Distillation through Dataset Dictionary Learning},
pdfsubject={cs.LG,stat.ML,cs.AI},
pdfauthor={Eduardo Fernandes Montesuma,Fred Ngolè Mboula,Antoine Souloumiac},
pdfkeywords={Optimal Transport,Dataset Distillation,Domain Adaptation,Dictionary Learning},
}

\newacronym{ot}{OT}{Optimal Transport}
\newacronym{erm}{ERM}{Empirical Risk Minimization}
\newacronym{ml}{ML}{Machine Learning}
\newacronym{dil}{DiL}{Dictionary Learning}
\newacronym{dd}{DD}{Dataset Distillation}
\newacronym{dc}{DC}{Dataset Condensation}
\newacronym{dm}{DM}{Distribution Matching}
\newacronym{mmd}{MMD}{Maximum Mean Discrepancy}
\newacronym{dadil}{DaDiL}{Dataset Dictionary Learning}
\newacronym{spc}{SPC}{Samples Per Class}
\newacronym{ipm}{IPM}{Integral Probability Metric}
\newacronym{svm}{SVM}{Support Vector Machine}
\newacronym{da}{DA}{Domain Adaptation}
\newacronym{msda}{MSDA}{Multi-Source Domain Adaptation}
\newacronym{cstr}{CSTR}{Continuously Stirred Tank Reactor}
\newacronym{tep}{TEP}{Tennessee Eastmann Process}
\newacronym{har}{HAR}{Human Activity Recognition}
\newacronym{wbt}{WBT}{Wasserstein Barycenter Transport}
\newacronym{co}{CO}{Caltech-Office 10}
\newacronym{cwru}{CWRU}{Case Western Reserve University}
\newacronym{sota}{SOTA}{State-of-the-Art}

\tikzstyle{trapezium}=[fill=white, draw=black, shape=trapezium, rotate=-90, minimum height=1cm]
\tikzstyle{lossbox}=[fill={rgb,255: red,202; green,206; blue,255}, draw=black, shape=rectangle, minimum height=1.2cm, minimum width=1cm, align=center]
\tikzstyle{clfbox}=[fill=white, draw=black, shape=rectangle, minimum width=1cm, minimum height=1cm]
\tikzstyle{new style 2}=[fill=white, draw=black, shape=rectangle, align=center]
\tikzstyle{domainbox}=[fill=white, draw=black, shape=rectangle, minimum width=3cm, align=center]
\tikzstyle{longbox}=[fill=white, draw=black, shape=rectangle, minimum height=4cm, minimum width=1.2cm, align=center]
\tikzstyle{rotatednode}=[rotate=90]
\tikzstyle{circularnode}=[fill={rgb,255: red,204; green,204; blue,204}, draw=black, shape=circle]
\tikzstyle{blue_circle}=[fill={rgb,255: red,0; green,80; blue,104}, draw=none, shape=circle, minimum width=0.5cm]
\tikzstyle{orangecircle1}=[fill={rgb,255: red,231; green,111; blue,81}, draw=none, shape=circle, minimum width=0.5cm]
\tikzstyle{blue_square1}=[fill={rgb,255: red,0; green,80; blue,104}, draw=none, shape=rectangle, minimum width=0.5cm, minimum height=0.5cm]
\tikzstyle{blue_triangle1}=[fill={rgb,255: red,0; green,80; blue,104}, draw=none, shape=regular polygon, regular polygon sides=3]
\tikzstyle{widebox}=[fill=white, draw=black, shape=rectangle, minimum height=1.2cm, minimum width=10cm, align=center]
\tikzstyle{labeled domain}=[fill=none, draw={rgb,255: red,0; green,80; blue,104}, shape=circle, minimum width=1cm]
\tikzstyle{unlabeled domain}=[fill=none, draw={rgb,255: red,231; green,111; blue,81}, shape=circle, minimum width=1cm]

\tikzstyle{red edge}=[->, fill=none, draw={rgb,255: red,128; green,0; blue,0}]
\tikzstyle{blue edge}=[->, fill=none, draw={rgb,255: red,70; green,130; blue,180}]
\tikzstyle{green edge}=[->, fill=none, draw={rgb,255: red,44; green,160; blue,44}]
\tikzstyle{red dotted edge}=[->, dashed, fill=none, draw={rgb,255: red,128; green,0; blue,0}]
\tikzstyle{blue dotted edge}=[->, dashed, fill=none, draw={rgb,255: red,70; green,130; blue,180}]
\tikzstyle{green dotted edge}=[->, dashed, fill=none, draw={rgb,255: red,44; green,160; blue,44}]
\tikzstyle{red dotted line}=[-, fill=none, dashed, draw={rgb,255: red,128; green,0; blue,0}]
\tikzstyle{blue dotted line}=[-, fill=none, dashed, draw={rgb,255: red,70; green,130; blue,180}]
\tikzstyle{green dotted line}=[-, fill=none, dashed, draw={rgb,255: red,44; green,160; blue,44}]
\tikzstyle{black edge}=[->]
\tikzstyle{black dashed line}=[-, dashed]
\tikzstyle{thick black arrow}=[->, thick]
\tikzstyle{thick black edge}=[-, thick]

\renewcommand{\inf}[1]{\underset{#1}{\text{inf}}\,}
\renewcommand{\sup}[1]{\underset{#1}{\text{sup}}\,}

\newcommand{\iid}[0]{\overset{i.i.d.}{\sim}}
\newcommand{\arginf}[1]{\underset{#1}{\text{arginf}}\,}
\newcommand{\argmin}[1]{\underset{#1}{\text{argmin}}\,}

\definecolor{myorange}{rgb}{0.906,0.435,0.317}
\definecolor{myblue}{rgb}{0.0,0.314,0.408}

\theoremstyle{definition}

\newtheorem{definition}{Definition}

\begin{document}
\maketitle

\begin{abstract}
In this paper, we consider the intersection of two problems in machine learning: Multi-Source Domain Adaptation (MSDA) and Dataset Distillation (DD). On the one hand, the first considers adapting multiple heterogeneous labeled source domains to an unlabeled target domain. On the other hand, the second attacks the problem of synthesizing a small summary containing all the information about the datasets. We thus consider a new problem called MSDA-DD. To solve it, we adapt previous works in the MSDA literature, such as Wasserstein Barycenter Transport and Dataset Dictionary Learning, as well as DD method Distribution Matching. We thoroughly experiment with this novel problem on four benchmarks (Caltech-Office 10, Tennessee-Eastman Process, Continuous Stirred Tank Reactor, and Case Western Reserve University), where we show that, even with as little as 1 sample per class, one achieves state-of-the-art adaptation performance.
\end{abstract}


\section{Introduction}\label{sec:intro}

In modern \gls{ml} practice, researchers face the challenge of reasoning about large-scale, heterogeneous datasets. This situation is challenging, as intuitive geometric concepts lose sense in high dimensions, and the computational cost of processing large amounts of data is often prohibitive. As such,~\cite{wang2018dataset} proposed \gls{dd}, a novel field of \gls{ml} that seeks to synthesize a small dataset summary while retaining as much information as possible.

Nonetheless, current works in \gls{dd} still need to consider the heterogeneity present in datasets. An example of such a phenomenon occurs in \gls{msda}~\cite{crammer2008learning}, where datasets contain multiple domains that follow different but related probability distributions. In this context, previous algorithms~\cite{montesuma2021icassp,montesuma2021cvpr,montesuma2023learning} leverage Wasserstein barycenters~\cite{agueh2011barycenters} for performing \gls{msda}. As we argue in this paper, this mechanism can be used for distillation.

In this paper, we propose to bridge \gls{msda} and \gls{dd}, i.e., performing \gls{msda} while summarizing the target domain. We call this new problem \gls{msda}-\gls{dd}. To this end, we adapt previous \gls{msda} methods, such as \gls{wbt}~\cite{montesuma2021icassp,montesuma2021cvpr} and \gls{dadil}~\cite{montesuma2023learning}, and \gls{dd} method \gls{dm}~\cite{zhao2023dataset} to our setting. To the best of our knowledge, this is the first paper considering \gls{msda} and \gls{dd} simultaneously.

In the following, section~\ref{sec:related_work} discusses previous work on \gls{dd} and \gls{msda}. Section~\ref{sec:proposed_method} presents our methodology. Section~\ref{sec:experiments} shows and discusses our experiments. Finally, section~\ref{sec:conclusion} concludes this paper.

\section{Related Work}\label{sec:related_work}

\gls{dd} is a novel problem in \gls{ml}, founded by~\cite{wang2018dataset}, whose goal is to synthesize a small set of samples, which retain the information of the whole original dataset. So far, most techniques focus on distilling single datasets to train neural networks~\cite{zhao2021dataset,lee2022dataset}. In this sense,~\cite{zhao2023dataset} proposed a method where data summaries are synthesized by \emph{matching probability distributions}. Nonetheless, none of these works consider the case of distillation of \emph{closely related datasets} characterized by a shift in their probability distributions. In the following, we explore how to leverage the similarities in similar datasets to enhance the power of data summaries, i.e., being capable of synthesizing a few effective samples for \gls{da}.

In parallel, \gls{msda}~\cite{crammer2008learning} proposes adapting multiple domains with heterogeneous but related probability distributions towards an unlabeled target domain. This problem was previously explored by~\cite{montesuma2021icassp,montesuma2021cvpr} and~\cite{montesuma2023learning} through \emph{Wasserstein barycenters}~\cite{agueh2011barycenters}, i.e., calculating an equidistant distribution of sources in a Wasserstein space. An essential feature of empirical, free-support Wasserstein barycenters, as proposed by~\cite{montesuma2023learning}, is that the number of samples in their support is a hyper-parameter. As we explore in this work, we can leverage this property for \gls{msda} and \gls{dd}.

\section{Methodology}\label{sec:proposed_method}

\subsection{Classification and Domain Adaptation}\label{sec:erm}

Classification is a sub-problem in supervised learning, formalized through the \gls{erm} principle. For a distribution $Q$, a ground-truth $h_{0}:\mathbb{R}^{d}\rightarrow\{1,\cdots,n_{c}\}$ and a loss function $\mathcal{L}$, let $\mathbf{x}_{i}^{(Q)} \iid Q$ and $y_{i}^{(Q)} = h_{0}(\mathbf{x}_{i}^{(Q)})$. A classifier $h \in \mathcal{H}$ is said to fit the data, if,
\begin{align*}
    \hat{h} &= \argmin{h\in\mathcal{H}}\hat{\mathcal{R}}_{Q}(h) = \dfrac{1}{n}\sum_{i=1}^{n}\mathcal{L}(h(\mathbf{x}_{i}^{(Q)}),y_{i}^{(Q)}).
\end{align*}
where $\hat{\mathcal{R}}_{Q}$ is the empirical approximation for the \emph{true risk} $\mathcal{R}_{Q}$. This framework assumes that train and test data follow a single distribution $Q$. \gls{da} relaxes this assumption, by defining a domain as a pair $(\mathcal{X},Q(X))$ of a feature space and a distribution. Assuming $\mathcal{X} = \mathbb{R}^{d}$, domains differ when there is a shift $Q_{S}(X) \neq Q_{T}(X)$ in the distributions. As a consequence, classifiers fit with data from $Q_{S}$ may not be appropriate for data from $Q_{T}$.

\noindent\textbf{Problem Description.} \gls{msda} presupposes $\mathcal{Q}_{S} = \{Q_{S_{\ell}}\}_{\ell=1}^{N_{S}}$ of source distributions s.t. $Q_{S_{i}} \neq Q_{S_{j}}$, $\forall i \neq j$, and $Q_{T}$ s.t. $Q_{T} \neq Q_{S_{i}}$, $\forall i$. Even though $Q_{S_{\ell}}$ and $Q_{T}$ are not available, we have access to $(\mathbf{x}_{i}^{(Q_{S_{\ell}})}, y_{i}^{(Q_{\ell})}) \iid Q_{S_{\ell}}$ and $\mathbf{x}_{j}^{(Q_{T})} \iid Q_{T}$, i.e., the target domain is \emph{unlabeled}. We use labeled samples from sources and unlabeled samples from the target for creating a summary of labeled samples in the target domain.

\subsection{Metric and Barycenters of Probability Distributions}\label{sec:ot_ipms}

In this section we introduce the Wasserstein~\cite{villani2009optimal} and the \gls{mmd}~\cite{gretton2012kernel} distances, an \gls{ot}-based and a kernel-based metric, respectively. We refer readers to~\cite{peyre2017computational,montesuma2023learning} for a broader context of \gls{ot}, and~\cite{sriperumbudur2012on} for a review on metrics between distributions.

On the one hand, under a linear kernel, the \gls{mmd} is defined by,
\begin{equation}
    \text{MMD}(P,Q) = \lVert \mu^{(P)} - \mu^{(Q)} \rVert_{2},\label{eq:mmd}
\end{equation}
where $\mu^{(P)} = \int x dP$. On the other hand, let $\Pi(P, Q)$ denote the set of distributions with marginals $P$ and $Q$. Then,
\begin{equation}
    W_{c}(P, Q) = \arginf{\pi \in \Pi(P, Q)}\int c(\mathbf{x}^{(P)},\mathbf{x}^{(Q)})d\pi,\label{eq:wasserstein_dist}
\end{equation}
where $c$ is called \emph{ground-cost} (e.g., $W_{2}$, where $c$ is the Euclidean distance). In many \gls{ml} applications, $P$ and $Q$ are unknown. One then resorts to \emph{empirical approximations} via samples $\mathbf{x}_{i}^{(P)} \iid P$ and $\mathbf{x}_{j}^{(Q)} \iid Q$, so that,
\begin{align}
    \hat{P}(\mathbf{x}) &= \dfrac{1}{n}\sum_{i=1}^{n}\delta(\mathbf{x}-\mathbf{x}_{i}^{(P)}),\label{eq:empirical_approx}
\end{align}
where $\delta$ is the Dirac delta function and $\mathbf{X}^{(P)} \in \mathbb{R}^{n \times d}$ is the support of $\hat{P}$. Under such approximation, one can calculate MMD$(\hat{P},\hat{Q})$ and $W_{c}(\hat{P},\hat{Q})$ by plugging-in eq.~\ref{eq:empirical_approx} into~\ref{eq:mmd} and~\ref{eq:wasserstein_dist}.

A particularly useful application of \gls{ot} is defining barycenters of distributions. This concept allows for the aggregation of a family of distributions, and was first introduced by~\cite{agueh2011barycenters},
\begin{definition}\label{def:wbary}
For distributions $\mathcal{P} = \{P_{k}\}_{k=1}^{K}$ and weights $\alpha \in \Delta_{K}$, the Wasserstein barycenter is a solution to,
\begin{align}
    B^{\star} = \mathcal{B}(\alpha;\mathcal{P}) = \inf{B}\sum_{k=1}^{K}\alpha_{k}W_{c}(P_{k}, B).\label{eq:true_bary}
\end{align}
\end{definition}

Wasserstein barycenters can be calculated for empirical distributions~\cite{cuturi2014fast,montesuma2021icassp,montesuma2021cvpr}, in which case it is empirical as well. Henceforth we use~\cite[Algorithm 1]{montesuma2023learning} for calculating $\mathcal{B}(\alpha;\mathcal{P})$. Wasserstein barycenters are interesting for synthesizing small summaries of distributions, since they have a free number of samples in their support.

\subsection{Dataset Distillation through Distribution Matching}\label{sec:dm}

\gls{dd}~\cite{wang2018dataset} seeks to represent a dataset $\{\mathbf{x}_{i}^{(Q)}, y_{i}^{(Q)}\}_{i=1}^{n}$ through a new set of \emph{synthetic} samples $\{\mathbf{x}_{j}^{(P)}, y_{j}^{(P)}\}_{j=1}^{m}$, also called data summary of $Q$, such that $m \ll n$. In mathematical terms, \gls{dd} seeks for a summary $\hat{P}$ s.t. for $\epsilon > 0$, and $h_{0}$~\cite{sachdeva2023data},
\begin{align*}
    \sup{\mathbf{x} \sim Q } |\mathcal{L}(h_{P}(\mathbf{x}), h_{0}(\mathbf{x})) - \mathcal{L}(h_{Q}(\mathbf{x}), h_{0}(\mathbf{x})) | \leq \epsilon,
\end{align*}
that is, when trained on the summary $\hat{P}$, a classifier has similar generalization performance (on $Q$) than when trained with $\hat{Q}$. Among the existing methods for \gls{dd},~\cite{zhao2023dataset} proposed building $\hat{P}$ through \gls{dm}. Let $d$ denote a notion of discrepancy between probability distributions (e.g. $W_{2}$). \gls{dm} seeks the summary $\hat{P}$ closest to $\hat{Q}$ \emph{in distribution},
\begin{equation}
    \hat{P}^{\star} = \argmin{\mathbf{x}_{1}^{(P)}, \cdots, \mathbf{x}_{m}^{(P)}}d(\hat{P},\hat{Q}),\label{eq:dm}
\end{equation}
where $m = \text{SPC} \times n_{c}$, for \gls{spc} and a number of classes $n_{c}$. In~\cite{zhao2023dataset}, the authors use a \gls{mmd}-like metric between feature-label joint distributions,
\begin{equation}
    \text{MMD}_{c}(\hat{P},\hat{Q}) = \sum_{c=1}^{n_{c}}\lVert 
\mu_{c}^{(P)} - \mu_{c}^{(Q)} \rVert_{2}^{2},\label{eq:ipm_class}
\end{equation}
where $\mu_{c}^{(P)} \in \mathbb{R}^{d}$ (resp. $Q$) is the mean vector of samples $\mathbf{x}_{i}^{(P)}$ belonging to class $c$.

\subsection{MSDA through Dataset Distillation}\label{sec:multi-dd}

An advantage of free-support Wasserstein barycenters~\cite[Algorithm 2]{cuturi2014fast}, is that the support $\mathbf{X}^{(B)}$ of $\hat{B}$ has a free number of samples. This property was previously treated as a hyper-parameter in \gls{msda} works~\cite{montesuma2023learning}. In this work, we investigate this feature through the lens of \gls{dd}, as it can be straightforwardly used to compress domains in \gls{msda}. In what follows, we describe 3 adaptations of previously proposed algorithms for \gls{msda}-\gls{dm}.

\begin{figure}
    \centering
    \begin{subfigure}{0.45\linewidth}
        \centering
        \resizebox{\linewidth}{!}{\begin{tikzpicture}
	\begin{pgfonlayer}{nodelayer}
		\node [style=none] (1) at (1.5, 2) {\color{myblue}\faDatabase};
		\node [style=none] (3) at (6.25, 6.75) {\color{myblue}\faDatabase};
		\node [style={blue_circle}] (4) at (6, 3) {};
		\node [style=none] (5) at (9.25, 1.5) {\color{myblue}\faDatabase};
		\node [style=none] (6) at (1.75, 1) {$\hat{Q}_{S_{2}}$};
		\node [style=none] (7) at (6.25, 7.5) {$\hat{Q}_{S_{1}}$};
		\node [style=none] (8) at (9.25, 0.75) {$\hat{Q}_{S_{3}}$};
		\node [style=none] (10) at (6, 2.25) {$\hat{P}$};
		\node [style=none] (12) at (11.5, 8) {};
		\node [style=none] (13) at (0.5, 8) {};
		\node [style=none] (14) at (0.5, 0) {};
		\node [style=none] (15) at (11.5, 0) {};
		\node [style=none] (16) at (8.25, 3.75) {\textcolor{myorange}{\faDatabase}};
		\node [style=none] (18) at (8.25, 4.75) {$\hat{Q}_{T}$};
	\end{pgfonlayer}
	\begin{pgfonlayer}{edgelayer}
		\draw [style=black dashed line] (1.center) to (4);
		\draw [style=black dashed line] (4) to (5.center);
		\draw [style=black dashed line] (4) to (3.center);
		\draw [style=black dashed line] (4) to (16.center);
	\end{pgfonlayer}
\end{tikzpicture}}
        \caption{\gls{msda}-\gls{dm} and \gls{wbt}}
        \label{fig:multi-dd-dm}
    \end{subfigure}\hfill
    \begin{subfigure}{0.45\linewidth}
        \centering
        \resizebox{\linewidth}{!}{\begin{tikzpicture}
	\begin{pgfonlayer}{nodelayer}
		\node [style={blue_square1}] (4) at (3, 1.25) {};
		\node [style=none] (5) at (1.75, 2) {\color{myblue}\faDatabase};
		\node [style={blue_square1}] (6) at (4.75, 4.25) {};
		\node [style=none] (7) at (3.25, 5.5) {\color{myblue}\faDatabase};
		\node [style={blue_square1}] (8) at (8.75, 1.25) {};
		\node [style=none] (9) at (7.25, 2) {\color{myblue}\faDatabase};
		\node [style=none] (10) at (11.5, -1.25) {$\hat{P}_{1}$};
		\node [style=none] (11) at (2.75, 9) {$\hat{P}_{3}$};
		\node [style=none] (12) at (0, -1.25) {$\hat{P}_{2}$};
		\node [style=none] (13) at (1.75, 3) {$\hat{Q}_{S_{2}}$};
		\node [style=none] (14) at (7.25, 3) {$\hat{Q}_{S_{1}}$};
		\node [style=none] (15) at (3.25, 6.25) {$\hat{Q}_{S_{3}}$};
		\node [style={blue_triangle1}] (16) at (2.75, 8) {};
		\node [style={blue_triangle1}] (17) at (0, -0.25) {};
		\node [style={blue_triangle1}] (18) at (11.5, -0.25) {};
		\node [style=none] (19) at (3, 0.25) {$\hat{B}_{2}$};
		\node [style=none] (20) at (8.75, 0.25) {$\hat{B}_{1}$};
		\node [style=none] (21) at (4.75, 3.5) {$\hat{B}_{3}$};
		\node [style=none] (22) at (4.25, 1.75) {\color{myorange}\faDatabase};
		\node [style={blue_square1}] (23) at (5.5, 1) {};
		\node [style=none] (24) at (5.5, 0.25) {$\hat{B}_{T}$};
		\node [style=none] (25) at (4.25, 2.5) {$\hat{Q}_{T}$};
	\end{pgfonlayer}
	\begin{pgfonlayer}{edgelayer}
		\draw [style=black dashed line, bend left=45] (7.center) to (6);
		\draw [style=black dashed line, bend left=45, looseness=1.25] (9.center) to (8);
		\draw [style=black dashed line, bend left=45] (5.center) to (4);
		\draw [style=black dashed line] (18) to (17);
		\draw [style=black dashed line] (17) to (16);
		\draw [style=black dashed line] (16) to (18);
		\draw [style=black dashed line, bend left=45] (22.center) to (23);
	\end{pgfonlayer}
\end{tikzpicture}}
        \caption{DaDiL}
        \label{fig:multi-dd-dadil}
    \end{subfigure}\hfill
    \caption{Conceptual illustration of \gls{msda}-\gls{dm} methods, where sources $\hat{Q}_{S_{\ell}}$ are \textcolor{myblue}{labeled}, and the target $\hat{Q}_{T}$ is \textcolor{myorange}{unlabeled}. The distillation is done from true datasets (\faDatabase) towards data summaries (blue circles). We denote \gls{dadil}'s atoms by triangles. While \gls{msda}-\gls{dm} and \gls{wbt} move the barycenter of true datasets towards the target domain, \gls{dadil} learns to express datasets as Wasserstein barycenters of atoms.}
    \label{fig:overview_methods}
\end{figure}
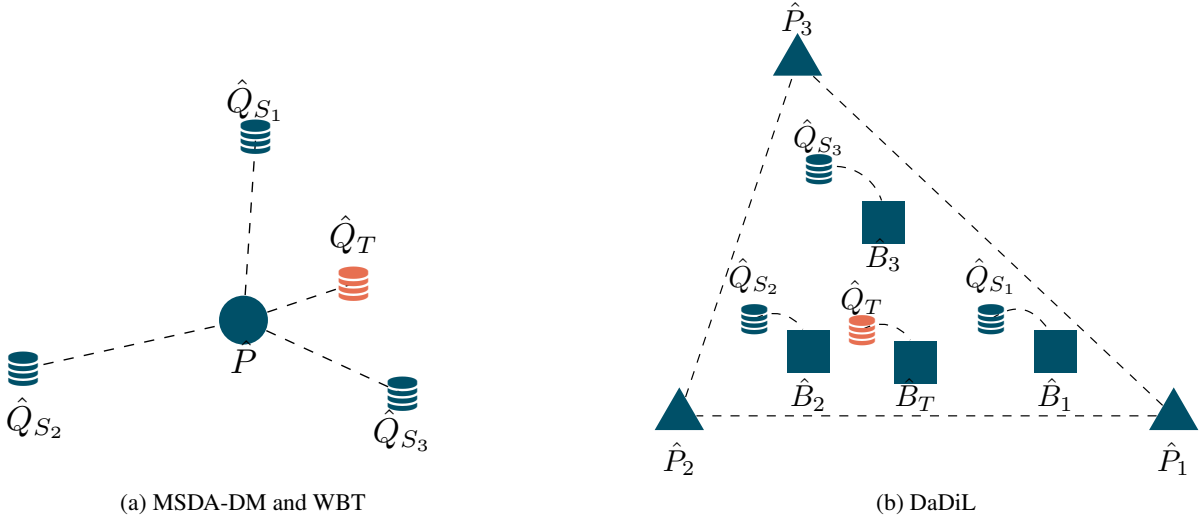

\noindent\textbf{Wasserstein Barycenter Transport.}~\cite{montesuma2021icassp,montesuma2021cvpr} previously proposed to use Wasserstein barycenters for \gls{msda}. Their approach minimizes the following objective,
\begin{align*}
    \hat{P} = \argmin{\{\mathbf{x}_{i}^{(P)},y_{i}^{(P)}\}_{i=1}^{m}}W_{2}(\hat{P},\hat{Q}_{T}) + \sum_{\ell=1}^{N_{S}}W_{c}(\hat{P},\hat{Q}_{S_{\ell}}),
\end{align*}
where $W_{2}$ and $W_{c}$ refer to Wasserstein distances with Euclidean ground-cost and $C_{ij} = \lVert \mathbf{x}_{i}^{(P)} - \mathbf{x}_{j}^{(Q)} \rVert_{2}^{2} + \beta \lVert \mathbf{y}_{i}^{(P)} - \mathbf{y}_{j}^{(Q)} \rVert_{2}^{2}$, for $\beta \gg \text{max}_{i,j}\lVert \mathbf{x}_{i}^{(P)} - \mathbf{x}_{j}^{(Q)} \rVert_{2}^{2}$, respectively. As such, \gls{wbt} first calculates a Wasserstein barycenter of sources, $\hat{B} = \mathcal{B}(\alpha;\mathcal{Q}_{S})$, with uniform weights $\alpha_{k} = N_{S}^{-1}$, then transports the barycenter to the target domain through a barycentric mapping~\cite[Eq. 14]{courty2017otda}. Hence, the target domain is compressed through $\hat{B}$.

\noindent\textbf{Dataset Condensation.} We adapt the framework of~\cite{zhao2023dataset} for \gls{msda}. Instead of minimizing the distance between the summary $\hat{P}$ and a single dataset $\hat{Q}$, the goal is to minimize,
\begin{align*}
    \hat{P} = \argmin{\{\mathbf{x}_{i}^{(P)}\}_{i=1}^{m}}\text{MMD}(\hat{P},\hat{Q}_{T})+\sum_{\ell=1}^{N_{S}}\text{MMD}_{c}(\hat{P},\hat{Q}_{S_{\ell}}).
\end{align*}

As we show conceptually in Fig.~\ref{fig:multi-dd-dm}, \gls{msda}-\gls{dm} and \gls{wbt} are conceptually close, in which they move the barycenter of labeled distributions towards the target, using the MMD, and $W_{2}$ respectively.

\noindent\textbf{Dataset Dictionary Learning.} In~\cite{montesuma2023learning}, authors proposed doing \gls{msda} through \gls{dil}. As such, \gls{dadil} learns a set of atoms $\mathcal{P} = \{\hat{P}_{k}\}_{k=1}^{K}$ and barycentric coordinates $\mathcal{A} = \{\alpha_{\ell}\}_{\ell=1}^{N_{S}+1}$, $\alpha_{\ell} \in \Delta_{K}$ and $\alpha_{T} := \alpha_{N_{S}+1}$. Let $\hat{B}_{\ell} = \mathcal{B}(\alpha_{\ell};\mathcal{P})$,
\begin{align*}
    (\mathcal{P}^{\star},\mathcal{A}^{\star}) &= \argmin{\mathcal{P},\mathcal{A}} W_{2}(\hat{Q}_{T},\hat{B}_{T})+\sum_{\ell=1}^{N_{S}}W_{c}(\hat{Q}_{\ell},\hat{B}_{\ell}).
\end{align*}

We illustrate \gls{dadil} conceptually in Fig.~\ref{fig:multi-dd-dadil}. Effectively, \gls{dadil} learns how to express each dataset $\hat{Q}_{\ell}$ as a Wasserstein barycenter of atoms, i.e., $\mathcal{B}(\alpha_{\ell};\mathcal{P})$. As a consequence, one can directly compress $\hat{Q}_{T}$ by calculating $\hat{B}_{T} = \mathcal{B}(\alpha_{T};\mathcal{P})$ with $n = SPC\times n_{c}$ points in its support.

\section{Experiments and Discussion}\label{sec:experiments}

In the following, we compare methods on 4 \gls{msda} benchmarks: (i) \gls{cstr}~\cite{pilario2018canonical,montesuma2022cross}, (ii) \gls{tep}~\cite{reinartz2021extended,montesuma2023multi}, (iii) \gls{cwru}\footnote{\url{https://engineering.case.edu/bearingdatacenter/download-data-file}} and (iv) \gls{co}~\cite{saenko2010adapting,griffin2007caltech}. While (i-iii) are fault diagnosis benchmarks, (iv) is a standard benchmark in visual \gls{da}. An overview is presented in table~\ref{tab:datasets}. The goal of tested algorithms is producing a small synthetic summary for the target domain.

For classification, we use a \gls{svm} over extracted features. For the \gls{cstr}, we use the norm of the power spectrum of each sensor data~\cite{montesuma2022cross}. For the \gls{tep}, \gls{cwru} and \gls{co} we use activations of neural networks, as in~\cite{montesuma2023multi} and~\cite{montesuma2023learning} respectively. All features are standardized to zero mean and unit variance.

\begin{table}[ht]
    \centering
    \caption{Overview of benchmarks used in our experiments.}
    \begin{tabular}{lllll}
         \toprule
         Benchmark & \# Samples & \# Domains & \# Classes & \# Features  \\
         \midrule
         CSTR & 2860 & 7 & 13 & 7\\
         TEP & 17289 & 6 & 29 & 128\\
         CWRU & 24000 & 3 & 10 & 256\\
         Caltech-Office 10 & 2533 & 4 & 10 & 4096\\
         \bottomrule
    \end{tabular}
    \label{tab:datasets}
\end{table}

In this setting, we compare 5 methods for \gls{msda}-\gls{dd}: (i) random sampling (source-only), (ii) random sampling (target-only), (iii) \gls{wbt}, (iv) \gls{msda}-\gls{dm} and (v) \gls{dadil}. On one hand, methods (iii-v) constitute our proposed adaptations for \gls{msda}-dd. On the other hand, (i,ii) are standard baselines in \gls{msda} and \gls{dd}. For (i), no adaptation is done towards the target, thus it is an intrinsically worst-case scenario. For (ii), there is no distribution shift, which characterizes it as a best case scenario. In both (i,ii), we randomly sample $n=SPC\times n_{c}$ samples from the overall data.

\begin{figure}[ht]
    \centering
    \includegraphics[width=\linewidth]{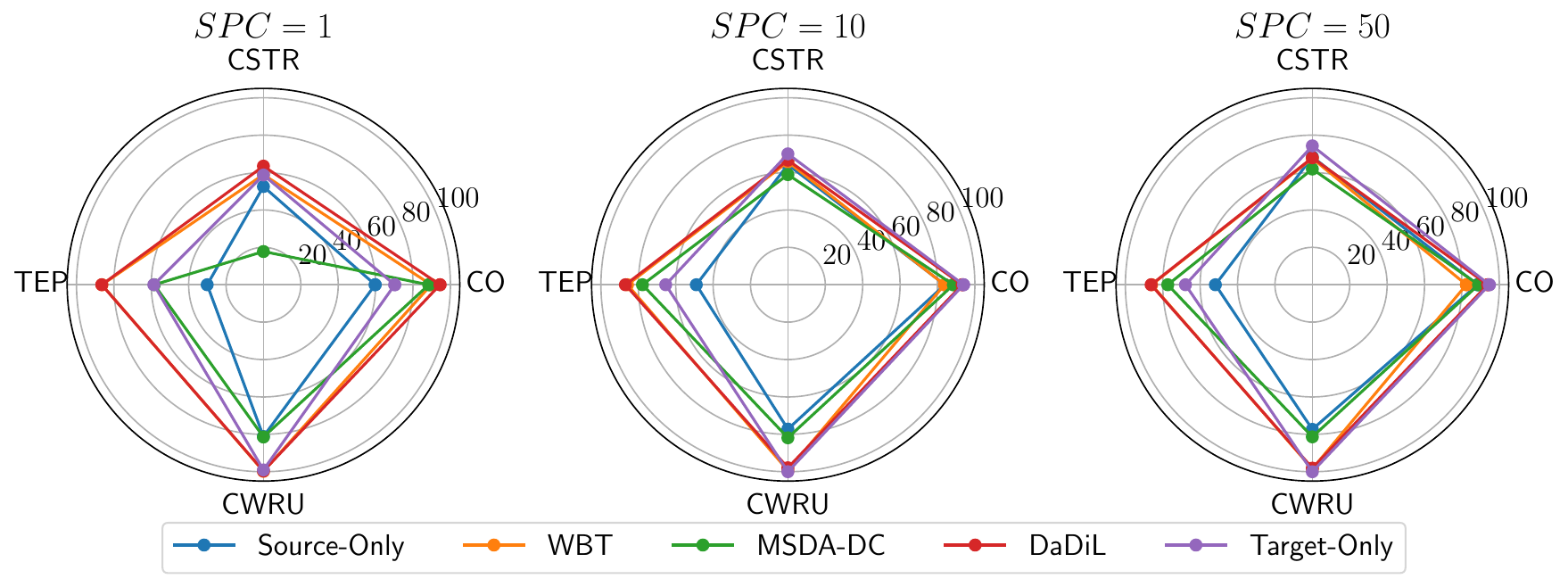}
    \caption{Global comparison of \gls{msda} methods in a distillation setting, for 1, 10 and 50 \gls{spc}.}
    \label{fig:overview}
\end{figure}

First, we present an overview of our results in Fig.~\ref{fig:overview}, as a function of \gls{spc}. Globally, \gls{dadil} and \gls{wbt} are largely superior to the baseline and \gls{msda}-\gls{dm}, especially in the fault diagnosis benchmarks. Surprisingly, the performance gap is more marked for small values of \gls{spc}, indicating that these methods are able to provide better generalization. This remark shows that \gls{msda} is possible, even with as little as 1 sample for each class in the target domain.

\begin{wrapfigure}{r}{0.5\linewidth}
\centering
\includegraphics[width=0.8\linewidth]{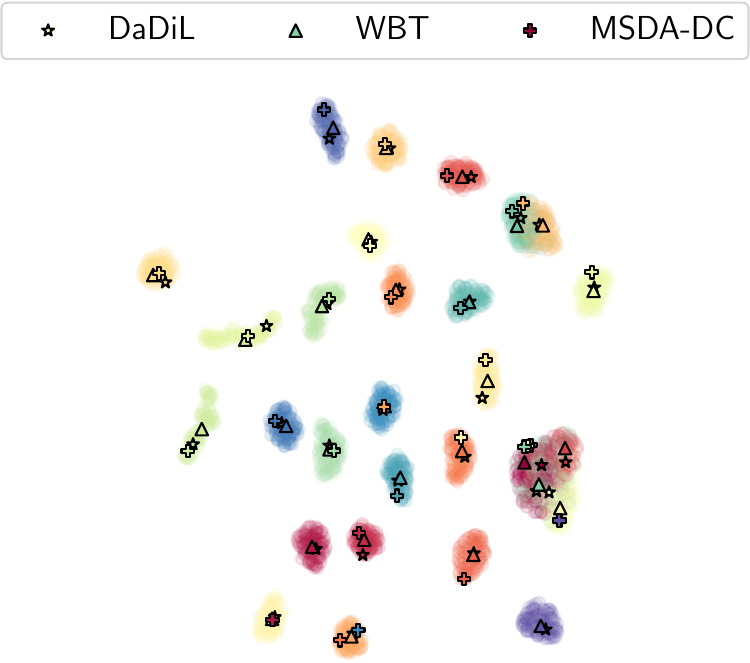}
\caption{UMAP projections of the \gls{tep} data for mode 1, where each color represent a different class.}
\vspace{-5pt}
\label{fig:umap_tep}
\end{wrapfigure}
Second, unlike \gls{dd}~\cite{wang2018dataset,zhao2023dataset}, for an increasing $SPC$, \gls{msda}-\gls{dd} methods do not converge to random sampling. Indeed, random sampling the source domain do not tackle the distributional shift problem. On the other hand one can expect random sampling the target domain to be optimal for large $SPC$ (e.g., \gls{cwru} and \gls{co} in Fig.~\ref{fig:overview}). Nonetheless, as demonstrated in Fig.~\ref{fig:overview}, \gls{msda}-\gls{dd} are \emph{sample efficient}, in the sense that they achieve high performance with as little as $SPC=1$, which represents 0.04\%, 0.16\%, 0.39\%, and 0.45\% of the overall number of samples in \gls{cwru}, \gls{tep}, \gls{co} and \gls{cstr} benchmarks.

Due to space constraints, we detail results only on the \gls{tep} benchmark. We explore how adaptation evolves for various values of \gls{spc} in the context of the 5 methods. Contrary to other benchmarks, on \gls{tep} performance remains stable over the range $SPC \in \{1,\cdots,50\}$. On one hand, \gls{wbt} and \gls{dadil} have nearly equivalent performance, which agrees with previous research on this benchmark~\cite{montesuma2023multi}. On the other hand, we are able to reach \gls{sota}, and to improve over the optimistic target-only scenario with only 1\% of target domain samples, or 0.16\% of the overall number of samples.

Next, we focus on the performance gap between \gls{wbt}, \gls{dadil} and \gls{msda}-\gls{dm}. In Fig.~\ref{fig:umap_tep} we show the UMAP~\cite{mcinnes2018umap} of target domain data, where the synthesized samples are highlighted. While \gls{wbt} and \gls{dadil} yield synthetic samples close to each other, \gls{msda}-\gls{dm} generates synthetic samples positioned in the wrong class cluster. This phenomenon indicates that the Wasserstein distance is a better candidate for \gls{dd}. Indeed, while the \emph{linear} \gls{mmd} is only able to match 1$^{st}$ order moments, the Wasserstein distance handles more complex distribution mismatch.

\begin{figure}[ht]
    \centering
    \includegraphics[width=\linewidth]{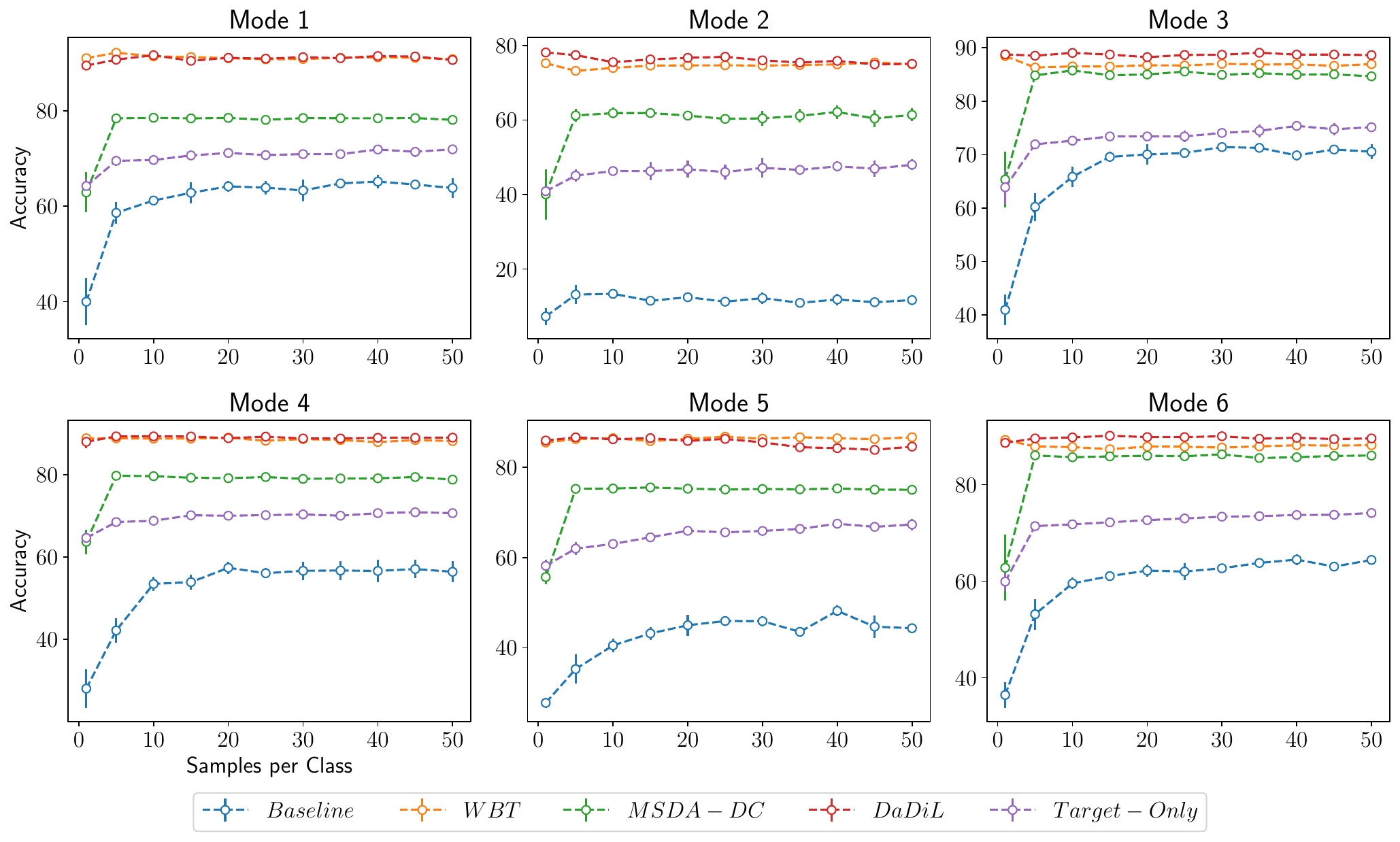}
    \caption{Classification accuracy as a function of \gls{spc}, for the 6 domains in the \gls{tep} benchmark. Error bars indicate 95\% confidence intervals.}
    \label{fig:da_results_tep}
\end{figure}

\section{Conclusion}\label{sec:conclusion}

In this paper, we bridge two fields of \gls{ml}: \gls{msda} and \gls{dd}. We propose a new problem, called \gls{msda}-\gls{dd}, were concurrently to \gls{msda} one also seeks to summarize the unlabeled target domain with labeled samples, while retaining as much information as possible. To that end, we adapt 3 \gls{sota} methods,~\cite{montesuma2021icassp,montesuma2021cvpr},\cite{montesuma2023learning}, and \cite{zhao2023dataset} to our setting. Data summaries generated by these methods capture knowledge from the multiple labeled source domains and the unlabeled target domain itself. We experiment extensively on 3 fault diagnosis benchmarks (\gls{cstr}, \gls{tep}, and \gls{cwru}) and 1 visual \gls{da} benchmark (Caltech-Office 10).

Our experiments show a series of intriguing results. First, we achieve \gls{sota} performance through \gls{wbt}~\cite{montesuma2021icassp,montesuma2021cvpr} and \gls{dadil}~\cite{montesuma2023learning} with only 1 sample per class. For instance, in the context of the \gls{tep} benchmark, this represents only $1\%$ of the samples in the target domain and $0.16\%$ of the overall number of samples. Second, unlike previous studies in standard \gls{dd}~\cite{wang2018dataset,zhao2023dataset}, \gls{msda} with \gls{dd} is not equivalent to random sampling when \gls{spc} is large. This remark is due to the distributional shift phenomenon involved in \gls{msda}.

Our work opens an interesting line of research, combining \gls{msda} and \gls{dd}. Future works include domain-incremental learning and considering label shifts between the different domains in \gls{msda}.

\bibliographystyle{apalike}
\bibliography{references}  






\end{document}